\pdfoutput=1

\documentclass[11pt]{article}

\usepackage[]{ACL2023}
\usepackage{times}
\usepackage{latexsym}
\usepackage{enumitem}
\usepackage[T1]{fontenc}

\usepackage[utf8]{inputenc}

\usepackage{microtype}

\usepackage{inconsolata}

\usepackage{color} 
\usepackage{soul} 
\usepackage{booktabs,colortbl}
\usepackage{times}
\usepackage{latexsym}
\usepackage{microtype}
\usepackage{amsmath}
\usepackage{amssymb}
\usepackage{multirow}
\usepackage{multicol}
\usepackage{bbding}
\usepackage{graphicx}
\usepackage{xfrac}
\usepackage{subfigure}
\usepackage{makecell}
\usepackage{cleveref}
\usepackage{arydshln}
\usepackage{color}
\usepackage[T1]{fontenc}

\usepackage[utf8]{inputenc}

\usepackage{microtype}

\usepackage{inconsolata}
\usepackage{tree-dvips}
\usepackage{qtree}
\usepackage{tikz}
\usetikzlibrary{trees}
\usepackage{CJKutf8}
\usepackage[linguistics]{forest}
\usepackage{enumitem}
\usepackage{tikz}
\usepackage{fontawesome5}
\usepackage{xcolor}
\usepackage{cleveref}
  {\list{}{\leftmargin=#1\rightmargin=#1}\item[]}%
  {\endlist}

\title{A Benchmark for Text Expansion: Datasets, Metrics, and Baselines}

  \author{Yi Chen$^{1}$, Haiyun Jiang$^{1*}$, Wei Bi$^{1}$, Rui Wang, Longyue Wang$^{1}$ \\
{\bf Shuming Shi$^{1}$, Ruifeng Xu$^{2}$}\thanks{\ \ Corresponding Authors.} \\
  $^{1}$Tencent AI Lab, Shenzhen, China \\
  $^{2}$Peng Cheng Laboratory, Shenzhen, China \\
  {\tt yichennlp@gmail.com} \\
  {\tt \{haiyunjiang,victoriabi,vinnylywang,shumingshi\}@tencent.com} \\
  {\tt xuruifeng.hitsz@gmail.com}
  }

\begin{document}
\begin{CJK*}{UTF8}{gbsn}
\maketitle
\begin{abstract}
This work presents a new task of Text Expansion (TE), which aims to insert fine-grained modifiers into proper locations of the plain text to concretize or vivify human writings. 
Different from existing insertion-based writing assistance tasks, TE requires the model to be more flexible in both locating and generation, and also more cautious in keeping basic semantics. 
We leverage four complementary approaches to construct a dataset with 12 million automatically generated instances and 2K human-annotated references for both English and Chinese. 
To facilitate automatic evaluation, we design various metrics from multiple perspectives. 
In particular, we propose Info-Gain to effectively measure the informativeness of expansions, which is an important quality dimension in TE. 
On top of a pre-trained text-infilling model, we build both pipelined and joint Locate\&Infill models, which demonstrate the superiority over the Text2Text baselines, especially in expansion informativeness. 
Experiments verify the feasibility of the TE task and point out potential directions for future research toward better automatic text expansion.
\end{abstract}

\section{Introduction}

\begin{table}[!t]
    \centering
    \small
    \begin{tabular}{ p{7.3cm} }
    \arrayrulecolor{blue!30}
    \toprule
    \rowcolor{blue!8}
    \textbf{Chinese Example} \\
    \midrule
      $X$: {\fontsize{8.5pt}{12pt}林丹在伦敦奥运会击败了李宗伟，获得了奥运冠军} \\
      (\textit{Lin Dan beat Lee Chong Wei at the London Olympics to win the Olympic title}) \\
      [0.5ex]\cdashline{1-1}\noalign{\vskip 0.8ex}

      $Y$:{\fontsize{8.5pt}{12pt}\colorbox[HTML]{ffe5ec}{而立之年的}\colorbox[HTML]{ffffff}{林丹在}\colorbox[HTML]{ffe5ec}{万众瞩目的}\colorbox[HTML]{ffffff}{伦敦奥运会}\colorbox[HTML]{fff4df}{一路}}\\
      {\tiny \hspace{0.8cm}\colorbox[HTML]{ffe5ec}{{[ADJP]}}\hspace{1.9cm}\colorbox[HTML]{ffe5ec}{ {[ADJP]}}}\\
      {\fontsize{8.5pt}{12pt}\colorbox[HTML]{fff4df}{过关斩将}\colorbox[HTML]{ffffff}{击败了}\colorbox[HTML]{ffe5ec}{实力强劲的}\colorbox[HTML]{ffffff}{李宗伟，}\colorbox[HTML]{ffe0ff}{如愿以偿地}} \\
      {\tiny\colorbox[HTML]{fff4df}{{[IDIOM]}}\hspace{1.95cm}\colorbox[HTML]{ffe5ec}{ {[ADJP]}} \hspace{2.15cm}\colorbox[HTML]{ffe0ff}{{[ADVP]}}}\\
      {\fontsize{8.5pt}{12pt}获得了\colorbox[HTML]{ffe5ec}{梦寐以求的}奥运冠军} \\
      {\tiny \hspace{1.33cm}\colorbox[HTML]{ffe5ec}{ {[ADJP]}}}\\
      (\textit{Lin Dan, in his 30s, went all the way to beat the powerful Lee Chong Wei at the highly anticipated London Olympics to win the coveted Olympic title}) \\
      \midrule
      \rowcolor{blue!8}
      {English Example}\\
      \midrule
   
      $X$: my favorite sport is basketball \\
      [0.5ex]\cdashline{1-1}\noalign{\vskip 0.8ex}
      

      $Y$:\colorbox[HTML]{e3f4ff}{{when it comes to sports ,}}\colorbox[HTML]{ffffff}{my}\colorbox[HTML]{ffe5ec}{{absolute}}\colorbox[HTML]{ffffff}{favorite sport}\\
      {\tiny \hspace{1.6cm} \colorbox[HTML]{e3f4ff}{[{SBAR}]}\hspace{2cm}\colorbox[HTML]{ffe5ec}{[{ADJP}]}} \\
      \colorbox[HTML]{dfffff}{{of all time}}\colorbox[HTML]{ffffff}{is basketball}\colorbox[HTML]{ffffdc}{{, and i 'm a huge fan .}} \\
      {\tiny \hspace{0.35cm}\colorbox[HTML]{dfffff}{{[PREP]}}\hspace{2.9cm}\colorbox[HTML]{ffffdc}{{[OTHERS]}}}\\
      
      \bottomrule
    \end{tabular}
    \caption{Examples of the Text Expansion task. $X$ and $Y$ denote the source and expanded texts, respectively. 
    The spans highlighted by different colors are expanded modifiers of various types indicated by subscripts, including \colorbox[HTML]{ffe5ec}{adjectives}, \colorbox[HTML]{ffe0ff}{adverbs}, \colorbox[HTML]{fff4df}{idioms}, \colorbox[HTML]{dfffff}{prepositions}, \colorbox[HTML]{e3f4ff}{subordinate clauses}, and \colorbox[HTML]{ffffdc}{other free-form modifiers}. 
    } 
    \label{tab:expansion_example}
\end{table}

In recent years, writing assistant systems \cite{
shi2023effidit} have made significant progress in various tasks. They can ``rewrite'' text to use different expressions \cite{bandel-etal-2022-quality,chen2022mcpg,shen2022evaluation} and correct grammar \cite{omelianchuk-etal-2020-gector}, as well as ``insert'' new content \cite{su2022a, ippolito-etal-2019-unsupervised, Coenen2021WordcraftAH} for inspiration and efficiency.

This paper introduces a new task called \textbf{T}ext \textbf{E}xpansion (TE) that aims to expand text by seamlessly inserting fine-grained modifiers, making it more concrete and vivid. 
Some examples are provided in Table~\ref{tab:expansion_example}. 
It is important to highlight the distinctions between TE and traditional query/document expansion in Information Retrieval (IR), the purpose of which is to augment queries or documents with relevant terms to address query-document mismatches without integrating the new terms seamlessly with the original text to form a coherent whole \cite{tang2017end, claveau2021neural, jeong2021unsupervised}.
Additionally, TE differs from other insertion-based writing assistant tasks in two key aspects:
(1) For application, it offers a novel functionality for users seeking to polish or enrich their writing in a nuanced way.
(2) For research, it requires a new formulation paradigm to accommodate flexible insertion locations and diverse modifier forms while maintaining semantic coherence, as exemplified in Table~\ref{tab:expansion_example}. 
More discussions on task differences can be seen in \S \ref{sec: related_work}.

To foster research on TE, we provide large-scale datasets, tailor evaluation metrics in various aspects, and develop baselines for comparison. 
We construct a training corpus of 12M parallel samples and a test set of 2K human-edited references for both Chinese and English, respectively. 
The training corpus is automatically constructed through several complementary approaches, which result in a more thorough coverage of various text-expansion pairs. 
To facilitate evaluation, we design a variety of automatic metrics. 
In particular, we propose Info-Gain for evaluating expansion informativeness. 
It is based on the inference perplexity from expanded modifiers to the source text and is highly consistent with human assessment.

We formulate the problem of TE, the goal of which includes both positioning and generating appropriate modifiers. 
We try two kinds of objectives, including Text2Text which treats TE as an end-to-end plain text generation task, and Locate\&Infill which explicitly locates the insertion places, and 
formulates modifier generation as a text-infilling problem. 
We find that handling the problem in a text-infilling manner is more suitable for insertion-based TE. 
Besides, using a joint framework for Locate\&Infill performs better in the English scenario, while the opposite is true for Chinese.

In summary, our main contributions include: 
\begin{itemize}[wide=0\parindent,noitemsep, topsep=0.5pt]
    \item We present a new task, Text Expansion, which aims to enrich a given text by inserting fine-grained modifiers 
    without modifying the core semantics.

    \item We leverage several complementary approaches to automatically construct a large corpus for both Chinese and English. 



    \item We design a number of automatic metrics to facilitate the evaluation process. Besides, we develop baseline systems based on different objectives and different model backbones. 
    
\end{itemize}

\section{Related Work}\label{sec: related_work}
\paragraph{Intelligent Writing Assistant}

A creative intelligent writing assistant is expected to enhance text by providing helpful ideas. It can perform various functions such as Text Completion \cite{su2022a}, Story Infilling \cite{ippolito-etal-2019-unsupervised}, Elaboration \cite{Coenen2021WordcraftAH}, etc. In these tasks, either the specific location for adding new content or the element to be elaborated is predetermined. However, in the proposed TE task, both of these aspects need to be resolved, and there is a higher requirement for maintaining semantic coherence.
A closely related task to TE is Writing Polishment with Simile (WPS) \cite{Zhang2020WritingPW}, which aims to refine texts by incorporating similes. WPS can be seen as a simplified version of TE, as it only focuses on a single insertion position. Similes are typically expressed through prepositional phrases (e.g., ``\textit{like ...}'') or subordinate clauses (e.g., ``\textit{as if ...}''), making them easily identifiable using our approach based on constituency tree pruning (see \S \ref{sec:constituency_pruning}). In contrast, our dataset for TE includes not only similes but also a wide range of other modifiers, posing a greater challenge than WPS.

\paragraph{Lexically Constrained Text Generation}


The proposed task, TE, is a type of Lexically Constrained Text Generation (LCTG), where specific words or phrases must be included in the output. Two approaches are commonly used: applying the constraint during decoding \cite{hokamp-liu-2017-lexically, post-vilar-2018-fast,hu-etal-2019-improved, lu-etal-2022-neurologic}, which can be computationally intensive, or training models to copy terminology \cite{dinu-etal-2019-training, ijcai2020p496, lee-etal-2021-improving}, reducing decoding latency. Another method is edit-based parallel refinement \cite{stern2019insertion, susanto-etal-2020-lexically, he2021xlentmcmc, he-2021-parallel}, which predicts editing operations and adjusts tokens accordingly.
Text Infilling (TI) is related to LCTG and involves predicting missing spans in a text, similar to the Cloze task\cite{taylor1953cloze}. TI has been successfully employed in pretraining tasks for large language models, such as BERT-style models for fixed-length infilling \cite{devlin-etal-2019-bert} and T5-style models for variable-length infilling \cite{raffel2020exploring,lewis-etal-2020-bart}.



\section{Automatic Data Construction}\label{sec:automatic_construction}
Given a source text $X = [x_1, ..., x_N]$, TE aims to expand $X$ to a longer text, e.g., $Y = [x_1, \underline{y_1, ..., y_j}, x_2, ..., x_N, \underline{y_{j+1}, ..., y_M}]$, by inserting proper modifiers (e.g., $[y_1, ..., y_j]$ and $[y_{j+1}, ..., y_M]$). 
The insertion locations can be the beginning, the end, or anywhere inside $X$.
The goal is to enrich and deepen the content of $X$ without altering its central idea, as shown in Table~\ref{tab:expansion_example}. 


To construct a corpus for TE, we follow two directions to derive the parallel counterpart from a plain text $C$.
One direction is ``$Y \Rightarrow X$'', which treats $C$ as the target expansion $Y$ and shorten it to obtain corresponding $X$ (\S \ref{sec:sentence_compression}-\ref{sec:entity_elaboration}). 
The other is ``$X \Rightarrow Y$'', which takes $C$ as the source $X$ and directly expand it to $Y$ (\S \ref{sec:mask_prediction}). We will analyze connections among these approaches later in \S\ref{sec: complementarity}. 

\subsection{$Y \Rightarrow X$ by Neural Sentence Compression}\label{sec:sentence_compression}
Intuitively, TE is similar to the reverse process of \textbf{T}ext \textbf{S}ummarization (TS). 
However, most paragraph or document-level TS datasets are not suitable for us.  
They usually construct a summary by truncating long sentences or paragraphs containing redundant information. 
Our goal is to expand a source text by inserting appropriate modifiers which are fine-grained and coherent to the core semantics, e.g., 
an adjective phrase or a short descriptive clause. 
Therefore, we only take advantage of existing datasets for sentence-level summarization, also known as \textbf{S}entence \textbf{C}ompression (SC), where both the given text and the summary are short texts. 
We merge all the available SC datasets \cite{Filippova2013OvercomingTL, Rush2015ANA} to train a neural Seq2Seq model, which is utilized to generate compressions for a large number of open-domain texts. 
The compressions and the corresponding given texts are taken as candidate source-expansion pairs $\{X,Y\}$ for TE. 
Since we restrict TE to expand texts only through insertion, all the tokens from $X$ should appear in $Y$ and their relative order should remain the same. 
We further filter out candidates that do not meet this criteria. 
Note that, since there lack of suitable SC datasets for Chinese, 
we only apply this approach to English texts. 

\subsection{$Y \Rightarrow X$ by Constituency Tree Pruning}\label{sec:constituency_pruning}
\begin{figure}[!t]
    \centering
    \includegraphics[scale=0.7]{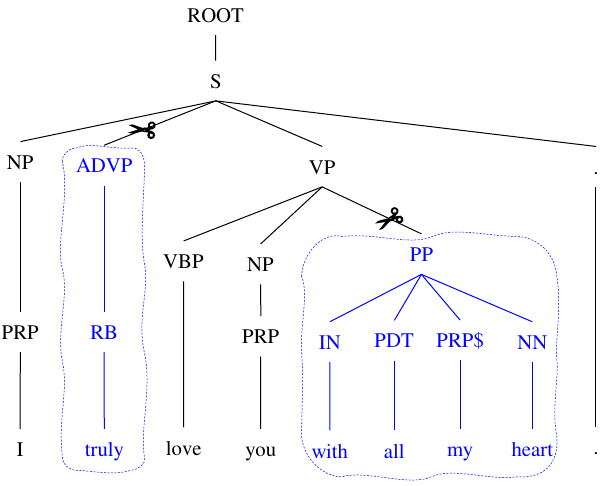}
    \caption{
    An example of constituency tree pruning (\ref{sec:constituency_pruning}). 
    The \textcolor{blue}{blue} parts refer to the pruned subtrees while the remaining tree indicates the extracted text skeleton.}
    \label{fig:en_constituency_tree_small}
\end{figure}

In this section, we propose a rule-based method for text skeleton extraction as a supplement to the neural approach in \S \ref{sec:sentence_compression}. 
We extract text skeleton by pruning inessential subtrees from the constituency tree parsed by Stanford CoreNLP\footnote{https://stanfordnlp.github.io/CoreNLP/}. As shown in Figure~\ref{fig:en_constituency_tree_small}, the inessential subtrees are mainly composed of modifier constituents that do not contribute to the primary content, e.g., the adverbial phrase  ``\textit{truly}'' (ADVP) and the prepositional phrase ``\textit{with all my heart}'' (PP). 
In this way, the brief skeleton and the whole text can serve as the source text $X$ and the target expansion $Y$ for TE. Detailed rules for both English and Chinese constituency tree pruning are presented in Appendix~\ref{sec: text_skeleton_extraction_rules}.

\subsection{$Y \Rightarrow X$ by IsA Relationship}\label{sec:entity_elaboration}
IsA relationship describes the semantic connection between
a subordinate term (i.e, hyponym) and the corresponding superordinate word or phrase (i.e., hypernym).
For example, ``pork'' is a kind of ``meat''. 
Hypernym is an important type of modifier for a hyponym in a text.
Following \cite{hearst-1992-automatic, roller-etal-2018-hearst}, we use the Hearst Patterns for hypernym detection. 
Some example patterns are presented in Appendix~\ref{sec:heart_patterns}.
Based on these patterns, we are able to extract the hypernym-based modifiers. 
For example, given $Y$ as ``\textit{we offer our buyers a wide range of {meat}, including {pork}, beef, and lamb}.''. 
We extract the modifier ``\textit{a wide range of {meat}, including}'', which consists of the hypernym phrase ``\textit{a wide range of meat}'' and the clue word ``\textit{including}''.
Then we derive $X$ as ``\textit{we offer our buyers {pork}, beef, and lamb.}'', where the hyponym phrase, ``\textit{pork, beef and lamb}'', is retained.
More example expansions can be found in Appendix~\ref{sec:heart_patterns}.


\subsection{$X \Rightarrow Y$ by Masked Modifier Prediction}\label{sec:mask_prediction}
We also take advantage of the pre-trained text-infilling model T5 \cite{Raffel2020ExploringTL} to directly produce expansion $Y$ for the source text $X$. 
Vanilla T5 is pre-trained by randomly masking some spans of a text and then recovering them. 
Following this paradigm, we can derive expansions by first inserting mask tokens into proper positions of $X$ and then using T5 to fill in these masked blanks. 

An example is shown in Table~\ref{tab:mmp}. 
To obtain appropriate modifiers, it is important to select suitable positions for insertion. 
Since nouns and verbs can be easily modified by some adjectives or adverbs, we  
select them as key modifiable anchors and insert mask placeholders adjacent to them. 
To select promising nouns, we can conduct entity typing \cite{qian2021fine,chen2021empirical,chen2022learning} on texts to obtain the entity information and further identify the subset to be modified.
To improve expansion quality, we further pre-train the text-infilling model by masking high-quality modifiers (including adjectives and adverbs) with a higher probability. 
For Chinese, idioms like ChengYu are also taken into account. 
The pre-training details are described in Appendix \ref{sec: masked_modifier_prediction_rules}.
Through further pretraining, the improved model is qualified to 
fill in the blanks with more elegant modifiers. 
In practice, to obtain more diverse expansions, the insertion places 
are not restricted to be around nouns and verbs. 
More details are presented in Appendix \ref{sec: masked_modifier_prediction_rules}.

\begin{table}[!t]
    \centering
    \small
    \begin{tabular}{ p{7.3cm} }
    \arrayrulecolor{blue!30}
    \toprule
    \rowcolor{blue!8}
    \textbf{Source Text: }
      {林丹在伦敦奥运会击败了李宗伟，获得了奥运冠军} 
      (\textit{Lin Dan beat Lee Chong Wei at the London Olympics to win the Olympic title}) \\
      \midrule
        \textbf{Modifiable Anchors: }
      \textcolor[HTML]{5b1da5}{林丹 ({\textit{Lin Dan}})} | \textcolor[HTML]{5b1da5}{伦敦奥运会 ({\textit{the London Olympics}})} | \textcolor[HTML]{c26462}{击败 ({\textit{defeated}})} | \textcolor[HTML]{5b1da5}{李宗伟 ({\textit{Lee Chong Wei}})} | \textcolor[HTML]{c26462}{获得 ({\textit{won}})} | \textcolor[HTML]{5b1da5}{奥运冠军 ({\textit{the Olympic champion}})}\\
      [1.0ex]\cdashline{1-1}\noalign{\vskip 1.0ex}
      \textbf{Infilling Input: }\texttt{\textcolor[HTML]{1a16cf}{<M1>}}{~林丹在~}\texttt{\textcolor[HTML]{1a16cf}{<M2>}}{~伦敦奥运会~}\texttt{\textcolor[HTML]{1a16cf}{<M3>}}{~击败了~}\texttt{\textcolor[HTML]{1a16cf}{<M4}>}{~李宗伟，\texttt{\textcolor[HTML]{1a16cf}{<M5>}~}获得了~}\texttt{\textcolor[HTML]{1a16cf}{<M6>}}{~奥运冠军}\\
      [1.0ex]\cdashline{1-1}\noalign{\vskip 1.0ex}
      \textbf{Infilled Expansion: }
    \textcolor[HTML]{1a16cf}{[而立之年的]$_1$~}林丹在\textcolor[HTML]{1a16cf}{~[万众瞩目的]$_2$~}伦敦奥运会\textcolor[HTML]{1a16cf}{~[一路过关斩将]$_3$~}击败了\textcolor[HTML]{1a16cf}{~[实力强劲的]$_4$~}李宗伟，\textcolor[HTML]{1a16cf}{~[如愿以偿地]$_5$~}获得了\textcolor[HTML]{1a16cf}{~[梦寐以求的]$_6$~}奥运冠军 (\textit{Lin Dan, in his 30s, went all the way to beat the powerful Lee Chong Wei at the highly anticipated London Olympics to win the coveted Olympic title})\\
      \bottomrule
    \end{tabular}
    \caption{An example for Masked Modifier Prediction. 
    The \textcolor[HTML]{9b5de5}{noun} and \textcolor[HTML]{c26462}{verb} anchors are highlighted in \textcolor[HTML]{9b5de5}{purple} and \textcolor[HTML]{c26462}{pink} respectively. 
    \texttt{\textcolor[HTML]{1a16cf}{<M*>}} indicates the i-th mask placeholder for infilling \textcolor[HTML]{1a16cf}{modifiers} adjacent to these anchors.} 
    \label{tab:mmp}
\end{table}

\subsection{Complementarity among Different Approaches}\label{sec: complementarity}


In \S \ref{sec:sentence_compression}$\sim $\S\ref{sec:mask_prediction}, we present four approaches based on Neural Sentence Compression (NSC), Constituency Tree Pruning (CTP),  IsA Relationship (IAR) and Masked Modifier Prediction (MMP) for automatic data construction. 
Table~\ref{tab:example_modifiers_by_different_approach} compares the example modifiers derived by different approaches. 

In the ``$Y\Rightarrow X$'' paradigm, we conclude that： 
\begin{itemize}[wide=0\parindent,noitemsep, topsep=0.5pt]
    \item NSC is able to extract some complicated patterns which are hard to be identified by CTP. But NSC datasets are not available in Chinese and are limited to the news domain in English. 
    \item  CTP is better at extracting fine-grained modifiers like adjective phrases which are easy to be covered by syntactic rules.
    \item IAR is responsible for extracting knowledge-rich hypernym-based modifiers, which is also a typical type of modifier but might be rarely covered by NSC and CTP.
\end{itemize}
This is the reason that we need NSC, CTP, and IAR to complement each other in this paradigm. 

Comparing the ``$Y\Rightarrow X$'' and ``$X\Rightarrow Y$'' paradigms, we conclude that:
\begin{itemize}[wide=0\parindent,noitemsep, topsep=0.5pt]
    \item ``$Y\Rightarrow X$''  limits the expansion quality 
 to some extent since $Y$ is a fixed given open-domain text, which usually contains only a few modifiers that are not always vivid or informative.  

    \item ``$X\Rightarrow Y$'' instead treats the open-domain texts as the source $X$ and directly expands them. In this way, we are able to insert as many high-quality modifiers as possible through MMP. 
    But since $Y$ is model-generated in this paradigm, sometimes it may be less fluent than human-written texts.
\end{itemize}
Therefore, both the ``$Y\Rightarrow X$'' and ``$X\Rightarrow Y$'' paradigms should be taken into account.

\begin{table}[!t]
    \centering
    \small
    \begin{tabular}{p{1.0cm} p{5.5cm}}
    \arrayrulecolor{blue!30}
    \toprule
    \rowcolor{blue!8}
    \multicolumn{2}{p{7.0cm}}{\textbf{Source Text:} we offer our buyers a wide range of meat , including pork , beef , and lamb} \\
    \midrule
    \multirow{8}{*}{$Y \Rightarrow X$}&\textbf{NSC:} we offer \textcolor[rgb]{0.3,0.3,0.7}{[\textit{our buyers}]} a wide range of meat \textcolor[rgb]{0.3,0.3,0.7}{[\textit{, including pork , beef , and lamb}]} \\
    [0.8ex]\cdashline{2-2}\noalign{\vskip 0.8ex}
    &\textbf{CTP:} we offer \textcolor[rgb]{0.3,0.3,0.7}{[\textit{our}]} buyers a \textcolor[rgb]{0.3,0.3,0.7}{[\textit{wide}]} range \textcolor[rgb]{0.3,0.3,0.7}{[\textit{of meat}} \textcolor[rgb]{0.3,0.3,0.7}{\textit{, including pork , beef , and lamb}]}\\
    [0.8ex]\cdashline{2-2}\noalign{\vskip 0.8ex}
    &\textbf{IAR:} we offer our buyers \textcolor[rgb]{0.3,0.3,0.7}{[\textit{a wide range of meat , including}]} pork , beef , and lamb\\
    \midrule
    \multirow{4}{*}{$X \Rightarrow Y$}&\textbf{MMP:} \textcolor[rgb]{0,0,1}{[\textit{we are pleased to inform that}]} we offer our \textcolor[rgb]{0,0,1}{[\textit{local}]} buyers a wide range of \textcolor[rgb]{0,0,1}{[\textit{ready-to-cook fresh}]} meat , including pork , beef , and lamb \textcolor[rgb]{0,0,1}{[\textit{at very affordable prices}]}\\
    \bottomrule
    \end{tabular}
    \caption{Example modifiers extracted or expanded by different approaches. The \textcolor[rgb]{0.3,0.3,0.7}{[\textit{gray blue}]} parts are the modifiers extracted from the source text while the \textcolor[rgb]{0,0,1}{[\textit{dark blue}]} parts are the modifiers expanded over the source text. }
    \label{tab:example_modifiers_by_different_approach}
\end{table}

\section{Data Annotation and Analysis}
\subsection{Data Statistics}
We use CommonCrawl\footnote{\url{https://commoncrawl.org/}} as the data source. 
Based on the four complementary approaches in \S \ref{sec:automatic_construction}, we further conduct data filtering (Appendix \ref{sec: data_filtering}) and construct a large-scale TE corpus with 12M pairs for both English and Chinese. Detailed statistics about data produced by different methods are listed in Appendix~\ref{sec: detailed_data_statistics}. 
In order to obtain high-quality references for testing, we randomly sample 2K pairs from the automatically constructed data and recruit annotators to manually refine them (\S \ref{sec: reference_refinement}). In the rest data, we randomly split 5K samples as the development set. 
For both English and Chinese, the train/dev/test splits are 12M/5K/2K.

\subsection{Quality Analysis}
To validate the quality of the automatically constructed data, before refining the 2K candidate references, we employ 3 well-educated annotators to assess their initial quality. 
The numbers of candidate pairs constructed by different approaches are proportional to those listed in Appendix~\ref{sec: detailed_data_statistics}. 
The annotators are required to answer two questions: (1) 
whether the expansion $Y$ is a fluent text, 
(2) 
whether the expanded modifiers in $Y$ are semantically coherent with $X$. 
A pair is labeled as eligible only if both questions are answered positively.


Table~\ref{tab:assessed_data_statistics} shows the quality of the assessed pairs. 
We observe that the proportion of eligible pairs for English and Chinese are {90.95\% and 87.80\%}, respectively. By average, 
$Y$ expands more than 10 tokens over $X$ at more than 3.5 positions. 
The results demonstrate that our approach for automatic data construction is reasonable for the TE task.

\subsection{Acquiring the Reference}\label{sec: reference_refinement}
We recruit 5 experienced annotators to carefully refine the automatically constructed 2K candidate pairs to acquire test references with high quality. 
For each candidate pair $\{X, Y\}$, 
the annotators are required to modify $Y$ according to the guidelines in Appendix \ref{sec: refinement_guide} to improve the expansion quality based on the criteria of various dimensions. 


\begin{table}[!t]
    \centering
    \small
    \begin{tabular}{l c c}
    \toprule
    {Corpus} & {English} & {Chinese} \\
    \midrule
    Num. of assessed & 2000 & 2000 \\
    Num. of eligible & 1819 & 1756 \\
    Prop. of eligible & 90.95\% & 87.80\% \\
    Av. length of $X$ & 13.87 & 19.15 \\
    Av. length of $Y$ & 24.71 & 36.04 \\
    Av. expanded length & 10.84 & 16.89 \\
    Av. num. of expansion pos. & 3.60 & 3.86 \\
    \bottomrule
    \end{tabular}
    \caption{Statistics of the assessed pairs.}
    \label{tab:assessed_data_statistics}
\end{table}



\section{Automatic Metrics}\label{sec: automatic_metrics}


We design several automatic metrics to measure the expansion quality from various perspectives as considered 
when annotating the reference (Appendix \ref{sec: refinement_guide}).
In particular, we propose Info-Gain to measure informativeness, which is an important quality dimension for TE.
Implementation details for these metrics are presented in Appendix~\ref{sec: implementation_details_metrics}.

\begin{itemize}[wide=0\parindent,noitemsep, topsep=0.5pt]
\item \textbf{Closeness to Reference (BLEU)}
We use BLEU \cite{papineni-etal-2002-bleu} to measure the textual similarity between generated results and references. 
Since the references are manually revised to improve the quality in each dimension (\S \ref{sec: reference_refinement}), 
a high BLEU similarity score might indicate the results are good in general. 
But note that, the TE task is open-ended while the number of references is limited, so a low BLEU score doesn't necessarily mean the generated expansion is bad. 


\item \textbf{Fidelity (Fidelity)}
is the success rate of the model to generate an expansion that keeps every token from the source text and maintains their order.

\item \textbf{Fluency (PPL)} \label{paragraph:language_fluency}
We use the output perplexity from a pre-trained language model, GPT2 \cite{radford2019language}
, to measure the fluency of expansions. 

\item \textbf{Fertility (N-Pos, Len)} 
We use the number (N-Pos) and the total length (Len) of inserted modifiers to show the surface-form richness of expansions. 



\item \textbf{Coherence (Nli-E) }\label{paragraph: semantic_fidelity} 
The semantic coherence between an expansion and the source text is analogous to textual entailment in Natural Language Inference (NLI)： a coherent expansion (premise) should semantically entail the source text (hypothesis). 
We use the entailment score (Nli-E) computed by a NLI model 
to capture this relationship. 
Appendix \ref{sec: incoherent_expansions} shows some incoherent expansions identified with low Nli-E scores.

\item \textbf{Informativeness (Info-Gain)}\label{parag: info_gain}
As shown in Table~\ref{tab:informative_examples},  a ``trivial expansion'' may only add some generic descriptions regardless of the content of the source text. 
However, besides ensuring semantic coherence, we also hope the expanded modifiers are creative and specific to the context as the ``informative expansion'' in Table~\ref{tab:informative_examples}. 
We find that non-trivial modifiers are usually informative for us to infer the context from the source text. 
Therefore, we propose a new metric, Info-Gain, to measure the informativeness of expansion results:
\begin{equation}
    \textit{Info-Gain} = \frac{\textit{Inherent-PPL}}{\textit{Infill-PPL}} \times \textit{Diff-Distinct}. 
\end{equation}
\end{itemize}
Infill-PPL is the perplexity to recover the source text based on the expanded modifiers. 
Inherent-PPL is the inherent perplexity of the source text. 
Diff-Distinct is a penalty coefficient. 

We use a pre-trained text-infilling model, T5, to estimate Infill-PPL and Inherent-PPL. Table~\ref{tab:informative_examples} shows an example ``input $\Rightarrow$ output'' format based on ``informative expansion''. 
The output perplexity of the text-infilling model reflects the likelihood to infill the context from source text on basis of the known modifiers. 
Lower perplexity indicates a higher probability. 
By dividing Inherent-PPL by Infill-PPL, we are able to assess the information gained by the expanded modifiers. 
To avoid favoring results that simply expand texts by repeating phrases from the source text, we introduce a penalty coefficient, \textit{Diff-Distinct}. 
It is calculated by averaging the proportions of unique n-grams ($n\in\{1, 2, 3, 4\}$) from the expanded modifiers that are not found in the source text. 

\begin{table}[!t]
    \centering
    \small
    \begin{tabular}{p{7cm}l}
    \arrayrulecolor{blue!30}
    \toprule
    \rowcolor{blue!8}
    \textbf{Source Text:} my favorite sport is basketball\\
    \midrule
     \textbf{Trivial Expansion:} \textcolor[rgb]{0.3,0.3,0.7}{[\textit{i 'm sure that}]} my favorite sport \textcolor[rgb]{0.3,0.3,0.7}{[\textit{, as you know ,}]} is basketball\\
     \midrule
     \textbf{Informative Expansion:} \textcolor[rgb]{0,0,1}{[\textit{besides tennis ,}]} my \textcolor[rgb]{0,0,1}{[\textit{personal}]} favorite sport \textcolor[rgb]{0,0,1}{[\textit{of all time}]} is basketball \textcolor[rgb]{0,0,1}{[\textit{, and i 'm a huge fan .}]} \\
     [1.0ex]\cdashline{1-1}\noalign{\vskip 1.0ex}
     \textbf{Infill-PPL:} \textcolor[rgb]{0,0,1}{besides tennis , }\texttt{<M1>} \textcolor[rgb]{0,0,1}{personal} \texttt{<M2>} \textcolor[rgb]{0,0,1}{of all time} \texttt{<M3>} \textcolor[rgb]{0,0,1}{, and i 'm a huge fan .}\\ $\Rightarrow$ 
     \vspace{0.1cm}
    \textcolor[rgb]{0,0,1}{\texttt{<M1>}} my \textcolor[rgb]{0,0,1}{\texttt{<M2>}} favorite sport \textcolor[rgb]{0,0,1}{\texttt{<M3>}} is basketball \textcolor[rgb]{0,0,1}{\texttt{<M4>}}\\

    \textbf{Inherent-PPL:} \texttt{<M1>} $\Rightarrow$ my favorite sport is basketball\\
     \bottomrule
    \end{tabular}
    \caption{Expansions with different degrees of informativeness, and the ``input $\Rightarrow$ output'' format for computing Infill-PPL and Inherent-PPL.
    In the input, \texttt{<M*>} are mask placeholders for the contextual spans from the source text. In the output, \textcolor[rgb]{0,0,1}{\texttt{<M*>}} are mask placeholders for the \textcolor[rgb]{0,0,1}{expanded modifiers} instead.
    }
    \label{tab:informative_examples}
\end{table}

\section{Baseline Models}
\begin{figure*}[!t]
    \centering
    \includegraphics[scale=0.465]{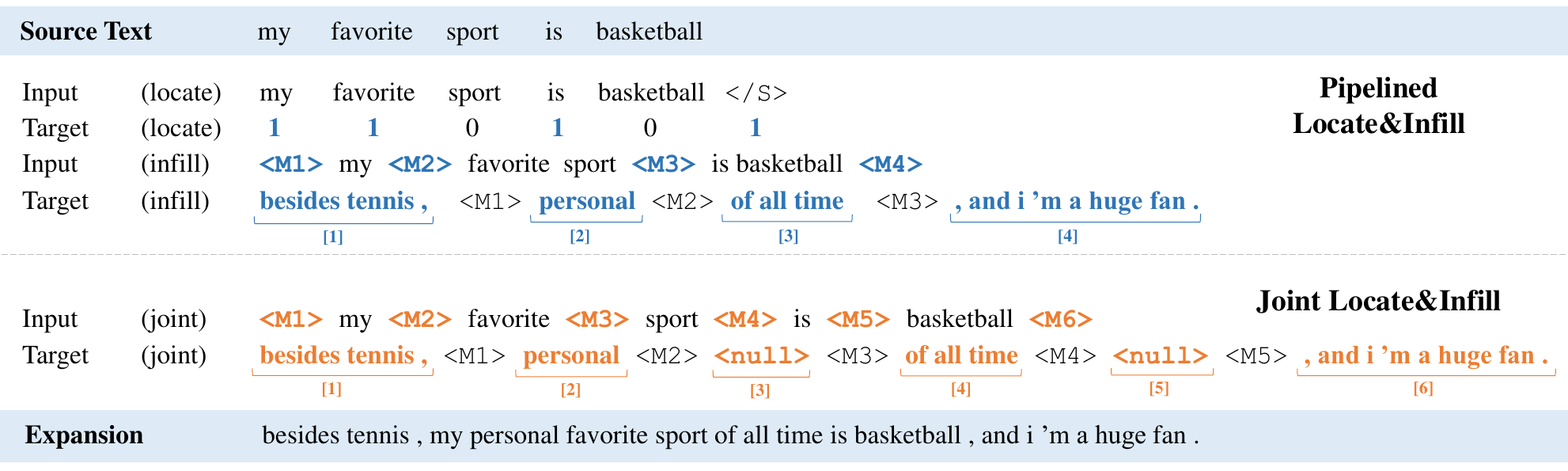}
    \caption{Schematics of the pipelined and joint Locate\&Infill frameworks.}
    \label{fig:piplined_and_joint_locate_infill}
\end{figure*}
There are two choices of objectives to tackle the TE task: (i) Text2Text, which is straightforward to simply take the source text as input and the expansion text as output without any formalization. (ii) Locate\&Infill, which is more controllable to explicitly locate the insertion places and expand modifiers in a text-infilling manner. 
Based on these two objectives, we implement several baseline. 
The implementation details are presented in Appendix~\ref{sec: implementation_details_baselines}. 

For the Text2Text objective, we choose both encoder-decoder based pre-trained models such as {T5} \cite{raffel2020exploring} and {BART} \cite{lewis-etal-2020-bart}, and the decoder-only pre-trained language model {GPT2} \cite{radford2019language}. 
For GPT2, the source and expansion texts are simply concatenated by a special token \texttt{[SEP]}. 
We denote these Text2Text baselines as: (1) \textbf{Text2Text-T5}, (2) \textbf{Text2Text-BART} and (3) \textbf{Text2Text-GPT2}.

For the Locate\&Infill objective, we implement both pipelined and joint frameworks. The schematics are shown in Fig~\ref{fig:piplined_and_joint_locate_infill}. 
\paragraph{(4) Pipelined Locate\&Infill:} 
We first fine-tune a token-level classifier based on BERT \cite{devlin-etal-2019-bert} to predict whether to insert a modifier before each token of the source text. A special token \texttt{</S>} is used to represent the end of the text. With labels from the location model, we insert mask tokens \texttt{<M*>} to corresponding positions and leverage a text-infilling model to fill in the variable-length modifiers. 
Since the input and target formats are the same as T5 \cite{raffel2020exploring}, we fine-tune the text-infilling model upon T5 to benefit from the pre-trained knowledge. 

\paragraph{(5) Joint Locate\&Infill:} 
We omit the separate location model and only rely on a text-infilling model to jointly locate and infill the modifiers. 
The text-infilling model is the same as used in the pipelined framework except that it introduces a special token \texttt{<null>} for insertion positioning. 
During inference, targeting each placeholder \texttt{<M*>} between every two adjacent tokens from the source text in the input, 
the model either predicts a span of modifier or just \texttt{<null>} when there's no suitable modifier to insert.

\section{Results and Analysis}

\begin{table*}[!t]
\small
    \centering
    \begin{tabular}{l c c c c c c c c c}
    \toprule
    Model & Len$\uparrow$  & N-Pos$\uparrow$ & PPL$\downarrow$ & Nli-E (\%)$\uparrow$ & Info-Gain$\uparrow$ & BLEU (\%)$\uparrow$ & Fidelity (\%)$\uparrow$ \\
    \midrule
    Text2Text-BART & 9.39 & 2.95 & 67.70 & \textbf{94.95} & 3.56 & 31.15 & 96.98 \\
    Text2Text-T5 & 9.91 & 3.13 & 70.69 & 94.65 & 3.75 & 30.81 & 94.99 \\
    Text2Text-GPT2 & 9.39 & 2.98 & 67.64 & 94.64 & 3.39 & 30.53& 96.28 \\
    [0.5ex]\cdashline{1-8}\noalign{\vskip 0.7ex}
    Pipelined Locate\&Infill & 9.67 & 2.68 & 69.00 & 94.63 & 3.70 & 33.25 & 99.95 \\
    \hspace{3em} w/~~ Oracle Location & 12.65 & \textbf{4.07} & \textbf{65.89} & 94.69 & \textbf{4.38} & \textbf{38.44} & \textbf{99.95} \\
    \hspace{3em} w/o Finetuning & 6.09 & 2.74 & 129.46 & 92.49 & 2.10 & 31.90 & 99.90 \\    
    Joint Locate\&Infill & 9.70 & 3.16 & 72.60 & 94.58 & 3.80 & 30.43 & \textbf{99.95} \\
    \hspace{3em} w/~~ Oracle Location & \textbf{13.76} & \textbf{4.07} & 70.60 & 94.19 & 4.26 & 36.80 & \textbf{99.75} \\
    \midrule
    Human Reference & 12.47 & {4.07} & 75.01 & {94.84} & {4.55} & - & {100.00} \\
    ChatGPT & {21.58} & 3.75 & {38.64} & {96.86} & - & 10.04 & 2.40 \\
    \bottomrule
    \end{tabular}
    \caption{The automatic evaluation results on the English TE task.}
    \label{tab: en_automatic_result}
\end{table*}

\begin{table*}[!t]
\small
    \centering
    \begin{tabular}{l c c c c c c c c c}
    \toprule
    Model & Len$\uparrow$  & N-Pos$\uparrow$ & PPL$\downarrow$ & Nli-E (\%)$\uparrow$ & Info-Gain$\uparrow$ & BLEU (\%)$\uparrow$ & Fidelity (\%)$\uparrow$ \\
    \midrule
     \noindent Text2Text-BART & 14.86 & 3.69 & 34.73 & 89.44 & 3.06 & 30.16 & 91.10 \\
    Text2Text-T5 & 15.34 & 3.84 & 33.90 & 89.26 & 3.19 & 31.06 & 93.85 \\
    Text2Text-GPT2 & 16.04 & 4.12 & 39.93 & 86.75 & 2.66 & 29.74 & 88.55 \\
    [0.5ex]\cdashline{1-8}\noalign{\vskip 0.7ex}
    Pipelined Locate\&Infill & \textbf{18.03} & \textbf{4.19} & \textbf{30.43} & 88.03 & 3.74 & {31.80} & \textbf{100.00} \\
    \hspace{3em} w/~~ Oracle Location & 17.43 & 4.14 & 30.68 & 89.15 & \textbf{3.85} & \textbf{38.49} & 99.90 \\
    \hspace{3em} w/o Finetuning &  11.19 & 4.17 & 48.19 & 82.76 & 2.58 & 28.63 & 99.60 \\
    Joint Locate\&Infill & 15.64 & 4.00 & 33.00 & \textbf{89.57} & {3.48} & 31.78 & \textbf{100.00} \\
    \hspace{3em} w/~~ Oracle Location & 17.60 & 4.14 & 33.13 & 88.68 & 3.59 & 37.59 & 99.40 \\
    \midrule
    Human Reference & {19.03} & 4.13 & 38.04 & 87.41 & {4.31} & - & {100.00} \\
    ChatGPT & {35.02} & 2.62 & {29.55} & {93.33} & - & 13.91 & 34.30\\
    \bottomrule
    \end{tabular}
    \caption{The automatic evaluation results on the Chinese TE task.}
    \label{tab: zh_automatic_result}
\end{table*}

\subsection{Overall Analysis}
\paragraph{Analysis of Baseline Models}
Table~\ref{tab: en_automatic_result} and Table~\ref{tab: zh_automatic_result} illustrate the automatic evaluation results on the English and Chines TE task, respectively. 
In general, the Locate\&Infill models are more suitable for TE, since the infilling paradigm guarantees a nearly 100\% fidelity score. 
Another obvious benefit of Locate\&Infill is that it yields a higher Info-Gain score than the Text2Text baselines. 
The model will be more focused on the adjacent context through infilling rather than continuation, which is helpful for generating context-specific and informative modifiers. 
We also observe that, in English, the Pipelined Locate\&Infill  model is slightly inferior to the Joint model on N-Pos and Info-Gain. 
While in Chinese, the opposite is true. 
This may be due to the difference in linguistic features. 
For Chinese, most modifiers are placed in front of the anchor elements (e.g., nouns and verbs), which makes it easier for deciding all the insertion locations at once before generating the whole expansion. 
However, in English, it depends more on the previously expanded context for insertion positioning since the places for inserting modifiers can be more flexible. 
Nevertheless, the Pipelined Locate\&Infill still achieves competitive results on English TE even with fewer modifiers inserted (indicated by the lower N-Pos score). 

\paragraph{Analysis of Human Reference}
It is interesting that the human-refined reference achieves slightly ``worse'' scores on PPL and Nli-E. 
It is because PPL might favor generic texts, which enjoy high frequencies when pre-training the language model. 
However, to create expansions with higher quality, human annotators tend to use more elegant modifiers, 
which are generally less frequent and thus result in slightly higher PPL scores. 
As for the Nli-E score, it might be influenced by the expansion length. 
Intuitively, the fewer changes the expansion makes over the source text, 
the easier it tends to be for the entailment reasoning, and the higher the Nli-E score is.
But making only minor changes is not what we expect. 
Therefore, we only encourage the PPL/Nli-E score to be within a normal range rather than relatively low/high. 

\paragraph{Analysis of ChatGPT}

To conduct a more comprehensive analysis, we also obtain expansions from ChatGPT using the OpenAI API\footnote{\url{https://platform.openai.com/playground?mode=chat\&model=gpt-3.5-turbo}} and report the results. Despite providing in-context examples and instructing ChatGPT to preserve the original text, it tends to rewrite the entire text, resulting in low fidelity scores. This indicates that ChatGPT is not proficient in generating constrained text. 
Besides, the Info-Gain score, derived using a text-infilling approach, requires expansions to meet fidelity conditions to differentiate between the added modifiers and the original text's contextual spans. However, only a small portion of ChatGPT's expansions faithfully replicate the source text, as indicated by the low fidelity scores. Consequently, the Info-Gain score for ChatGPT lacks statistical significance, and we do not report it in \Cref{tab: en_automatic_result,tab: zh_automatic_result}.
On other dimensions, ChatGPT excels in fluency (PPL) and coherence (Nli-E). This can be attributed to its extensive pre-training with billions of parameters and its intention to disregard generation constraints. 
Additionally, ChatGPT tends to produce longer texts (Len scores), which is a well-known characteristic of this model.

\subsection{Ablation Study}
\paragraph{Finetuning on the Constructed TE corpus}
In Table~\ref{tab: en_automatic_result} and Table~\ref{tab: zh_automatic_result}, we observe a distinct performance regression on Pipelined Locate\&Infill w/o (without) Finetuning when directly using the vanilla pre-trained model for infilling modifiers. 
Particularly, compared with Pipelined Locate\&Infill, the Info-Gain score decreases by around a third. 
Besides being less informative in content, the expansion results also become less fluent in expression (with a high PPL score) and less coherent in semantics (with a low Nli-E score). 
This demonstrates the effectiveness of the proposed data construction method since it brings significant gains in all metrics after fine-tuning the infilling model on the constructed TE corpus.

\paragraph{Controlling Insertion Locations}
The quality of expanded modifiers may be affected by the choice of  insertion locations. 
 We additionally report the performance of Pipelined and Joint Locate\&Infill w/ (with) Oracle Location when the models are forced to insert modifiers only to the oracle locations as the reference. 
With the pre-informed oracle locations, both the pipelined and the joint Locate\&Infill models are improved， particularly in the Info-Gain and BLEU scores. 
This reveals that increasing the accuracy of insertion positioning will benefit the model's performance and there is still room for improvement. 
It is also noticeable that the pipelined approach shows greater potential than the joint approach when the error accumulation from the positioning model is eliminated. 
Since in the pipelined approach, we do not make any change to the architecture of the pre-trained text-infilling model, it will benefit more from the built-in 
pre-trained knowledge for infilling modifiers. 
In real-world applications, the oracle insertion locations of reference are usually not available.
But it is promising that the Locate\&Infill framework allows users to customize where to insert the modifiers. 
With user intervention, the model will be adjusted to expand more appropriate modifiers that align better with human intentions.
Besides, some other technologies may contribute to refining position selection, e.g., keywords extraction \cite{song2023unsupervised} and relation extraction \cite{cheng2021question,chu2020insrl,jiang2019relation}.
\subsection{Human Assessment}
\begin{table}[!t]
    \centering
    \small
    \begin{tabular}{p{3.0cm} p{0.6cm}<{\centering} p{0.6cm}<{\centering} p{0.6cm}<{\centering} p{0.6cm}<{\centering}}
    \toprule
    Model & Flu. & Coh. & Info. & Eleg. \\
    \midrule
    Text2Text-BART & 97.25 & 98.50 & 68.00  & 53.25 \\
    Text2Text-T5 & 95.75 & 98.50 & 70.75 & 52.75 \\
    Text2Text-GPT2 & 93.00 & 98.00 & 55.25 & 47.25 \\
    Pipelined Locate\&Infill & 96.25 & 99.25 & \underline{86.74} & \underline{63.25} \\
    Joint Locate\&Infill & \underline{99.00} & \underline{99.50} & 79.75 & 56.50 \\
    \midrule
    Human Reference & \textbf{99.98} & \textbf{99.94} & \textbf{94.75} & \textbf{78.75} \\
    \bottomrule
    \end{tabular}
    \caption{Human evaluation results on expansion quality, depicted in four aspects: fluency, coherence, informativeness, and elegance. }
    \label{tab:human_evaluation_results}
\end{table}


We further conduct human evaluation to assess the expansion quality on four dimensions: fluency, coherence, informativeness, and elegance. 
The evaluation details are described in Appendix~\ref{sec: evaluation_guide}. 
In Table~\ref{tab:human_evaluation_results}, the results are consistent with the automatic evaluation results in Table~\ref{tab: zh_automatic_result}, where the Locate\&Infill baselines demonstrate a particular superiority over the Text2Text baselines. 
Besides, all the baselines are annotated with similar Flu. and Coh. scores that are close to the full scale. 
This indicates that fluency and coherence are not the primary challenges for the TE task. 
On the other hand, 
there shows a more distinct performance gap on Info. and Eleg. 
Even the best model scores in these two aspects are still inferior to the human-written reference by a large margin.  
This highlights the potential for TE models to be enhanced in the future to produce expansions that are both more informative and more elegant. 
Recall that we have proposed Info-Gain in \S \ref{parag: info_gain} 
to automatically measure the informativeness of expansions. 
We also calculate the Pearson and Spearman correlations between Info-Gain and human-judged Info. scores. 
The Pearson coefficient of 0.7769 and the Spearman coefficient of 0.7710 demonstrates the reliability of Info-Gain.

\section{Conclusion}
In this paper, we present a new task, \textbf{T}ext \textbf{E}xpansion (TE). 
The goal of TE is to insert proper modifiers to the source text
so as to make the resulting expansion more concrete and vivid,  
which is a useful function for the intelligent writing assistant.
We propose four complementary approaches to automatically construct a large corpus for both English and Chinese. 
The same paradigm is also applicable to studying TE in other languages. 
We establish several baselines based on different objectives and benchmark them on the human-annotated test dataset with our proposed evaluation metrics. 

\section*{Limitations}
\paragraph{Factual Error}
Currently, factual verification is not considered in TE. 
The models sometimes may generate plausible-sounding but nonsensical content as in many other open-ended text generation tasks. 
One possible solution is to divide the modifiers into factual and non-factual categories 
and allow the model to insert factual modifiers only when there is clear evidence for support. 


\paragraph{Lacking Informativeness}
Since generic modifiers enjoy higher frequency in the training corpus, they are also of higher probability to be generated during inference. 
Although we can ease this problem by adjusting the decoding procedure 
(e.g., controlling the temperature parameter to increase diversity), 
it may hurt the fluency meanwhile. 
Other possible methods may be to impose down-sampling upon the high-frequency modifiers during training or tailor modifiers toward certain objects based on an external knowledge base.

\paragraph{Improving Evaluation Metrics}
We have proposed several automatic metrics to evaluate the expansion quality from various aspects in \S \ref{sec: automatic_metrics}. 
However, a few more aspects should be considered in order to make a more comprehensive assessment, including the elegance and the factuality of expansions. 
Moreover, we now treat the expanded text as a whole for evaluation. 
It would be better if we could further assess the quality of each inserted modifier from a finer-grained perspective.


\clearpage
\bibliography{anthology,custom}
\bibliographystyle{acl_natbib}

\clearpage
\appendix
\begin{figure*}[!htbp]
    \centering
    \includegraphics[scale=0.7]{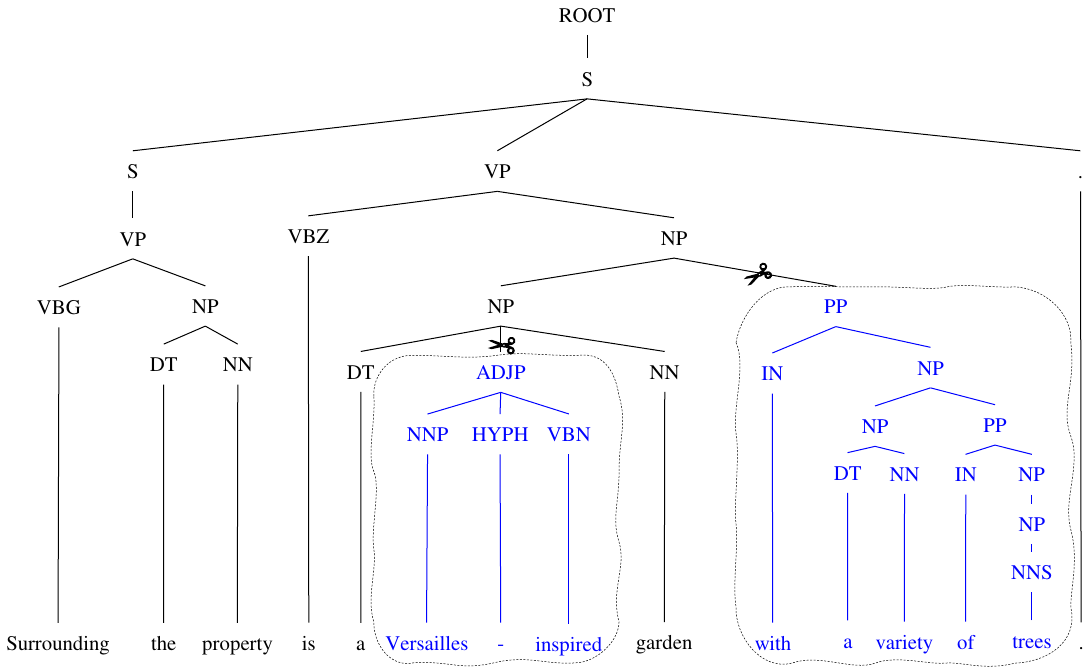}
    \caption{
    An example for English constituency tree pruning. 
    The \textcolor{blue}{blue} parts refer to the pruned subtrees while the remaining tree indicates the extracted text skeleton.}
    \label{fig:en_constituency_tree_example}
\end{figure*}
\begin{figure*}[!htbp]
    \centering
    \includegraphics[scale=0.7]{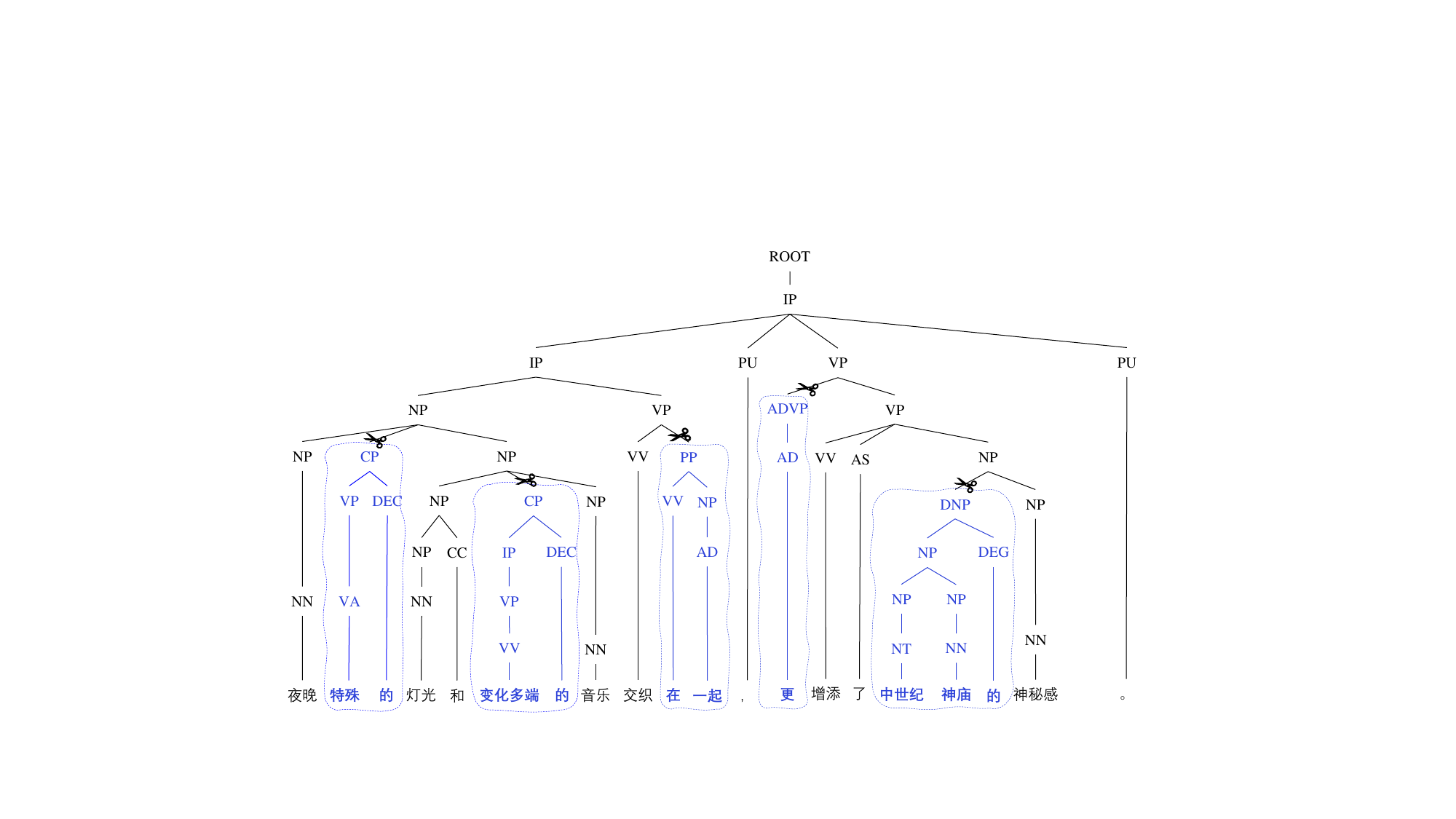}
    \caption{An example for Chinese constituency tree pruning.  
    The \textcolor{blue}{blue} parts refer to the pruned subtrees while the remaining tree indicates the extracted text skeleton.}
    \label{fig:zh_constituency_tree_example}
\end{figure*}
\section{Rules for Constituency Tree Pruning}\label{sec: text_skeleton_extraction_rules}
To extract the English text skeleton, a more complicated example is presented in Figure~\ref{fig:en_constituency_tree_example}. 
Starting from the ROOT node, we perform preorder traversal over the entire constituency tree. 
We decide whether to prune a subtree according to the constituent label of the top node. 
Some heuristic rules targeting nodes of different constituent labels are defined as follows: 
\begin{itemize}[itemsep=2pt,topsep=0pt,parsep=0pt]
    
    \item[(1)] \textbf{Default}: Process each child node recursively and 
    concatenate the results from all children.
    \item[(2)] \textbf{Leaf}: Return the surface string. 
    \item[(3)] \textbf{S}: Prune PPs and ADVPs.
    \item[(4)] \textbf{NP}: Prune noun-modifier child nodes such as ADJP, JJ, CD, PP, and all explanatory child nodes between bracket nodes `-LRB-' and `-RRB-'. 
    \item[(5)] \textbf{VP}: Prune PPs and verb modifiers such as ADVP. 
    \item[(6)] \textbf{ADJP}: Prune the adverb child nodes whose labels start with `RB'.
    \item[(7)] \textbf{SBAR}: Return null string if the label of first child node starts with `WH' which usually introduces a modifying clause w.r.t. adjective or adverbial clauses. 
\end{itemize}

For Chinese text skeleton extraction, as shown in Figure~\ref{fig:zh_constituency_tree_example}, we also adopt similar handcrafted rules:  
\begin{itemize}
    \item[(1) ] \textbf{Default}: Process each child node recursively while pruning DNPs, CPs， DVPs， ADVPs, QPs，LCPs and PPs. Concatenate the results from the remaining children.
    \item[(2)] \textbf{Leaf}: Return the surface string. 
\end{itemize}

For detailed meanings of these constituent tags, please refer to Stanford CoreNLP\footnote{https://stanfordnlp.github.io/CoreNLP/}.
Note, the subtrees are pruned selectively. 
For instance, we do not delete the nodes whose subtree contains more than 10 leaves 
to avoid truncating too much information. 

\section{Hearst Patterns for extracting Hypernym-based Modifiers}\label{sec:heart_patterns}
We present some example Heast patterns in Table~\ref{tab:hearst_patterns}. 
The hypernym-hyponym relationship is detected by some syntactic patterns combined with the clue words. 
The extracted modifier consists of both the hypernym phrase and the clue words, as highlighted in Table~\ref{tab:hearst_patterns}. Some example expansions resulted from this approach are listed in Table~\ref{tab:hypernym_expansions}.

\section{Details for Masked Modifier Prediction}\label{sec: masked_modifier_prediction_rules}
\paragraph{Pre-training}
Given a text, we first mask the high-quality modifiers as many as possible. Then we continue to mask the rest spans randomly until the masking rate reaches a threshold (25\%). The length of each masked span is limited to 1$\sim$10 tokens. 

\paragraph{Determining Insertion Places}
Given a text, we first randomly sample $3\sim5$ places to insert mask placeholders. 
The sampling process is repeated 5 times to produce several input sequences, where the mask placeholders are distributed differently. 
We select the resulting 5 infilled expansions according to their fluency.
Only the most fluent result with the lowest PPL (\S \ref{paragraph:language_fluency}) will be adopted.
To refine the sampling process, 
we reweight different candidate insertion places. 
We assign higher probabilities to the places around the noun and verb phrases.  
To prevent essential entities from being broken, we set the probabilities for places inside the named entities to 0.

\begin{table}[!t]
    \centering
    \small
    \begin{tabular}{l}
    \toprule
    \textbf{Pattern} \\
    \midrule
    NP$_\text{sub}$ \textcolor[rgb]{0,0,1}{[{is (a | an) (part | field | ...) of NP$_\text{sup}$ that}]} \\
    NP$_\text{sub}$ \textcolor[rgb]{0,0,1}{[{(and | or) (some | any) other NP$_\text{sup}$}]} \\
    NP$_\text{sub}$ \textcolor[rgb]{0,0,1}{[{(especially | particularly | ...) NP$_\text{sup}$}]} \\
    NP$_\text{sub}$ \textcolor[rgb]{0,0,1}{[{, (known | famous | ...) as NP$_\text{sup}$}]} \\
    \textcolor[rgb]{0,0,1}{[{NP$_\text{sup}$ (such as | including)}]} NP$_\text{sub}$ \\
    \textcolor[rgb]{0,0,1}{[such {NP$_\text{sup}$ as}]} NP$_\text{sub}$ \\
    \textcolor[rgb]{0,0,1}{[{NP$_\text{sup}$ (e.g. | i.e. | for instance | ...)}]} NP$_\text{sub}$ \\
    \textcolor[rgb]{0,0,1}{[{(known | famous | ...) as NP$_\text{sup}$ , }]} NP$_\text{sub}$  \\
    
    \bottomrule
    \end{tabular}
    \caption{Example Hearst patterns. 
    NP$_\text{sup}$ denotes the noun phrase containing the hypernym and NP$_\text{sub}$ denotes the hyponym phrase. The highlighted parts are the extracted hypernym-based modifiers.}
    \label{tab:hearst_patterns}
\end{table}

\begin{table}[!t]
    \centering
    \small
    \begin{tabular}{p{7.3cm}}
    \toprule
    \textbf{Hypernym-based Expansions}\\
    \midrule
    you may use baby food without added sugar or \textcolor[rgb]{0,0,1}{[\textit{starch such as}]} baby food chicken .\\
    \\
    you can use \textcolor[rgb]{0,0,1}{[\textit{the default wordpress search widget or any other}]} method that generates a standards-compliant search form .\\
    \\
    but \textcolor[rgb]{0,0,1}{[\textit{secondary pests, especially}]} mirids and mealy bugs , that are highly resistant to bt toxin , soon took its place .\\
    \\
    I really hate this elitist attitude whereby ordinary people cannot discuss \textcolor[rgb]{0,0,1}{[{\textit{such lofty topics as}}]} psychology and philosophy.\\
    \bottomrule
    \end{tabular}
    \caption{Some example expansions with \textcolor[rgb]{0,0,1}{[\textit{hypernym-based modifiers}]}.}
    \label{tab:hypernym_expansions}
\end{table}

\section{Data Filtering}\label{sec: data_filtering}
Through the approaches in \S \ref{sec:automatic_construction}, we obtain four sources of candidate text-expansion pairs for the TE task. 
The vanilla data may contain lots of noise. 
We further clean the large-scale 
corpus by some heuristic criteria. Pair $\{X, Y\}$ is discarded if any of the questions in Table~\ref{tab:filter_candidate} is answered positively.
Note, since the number of expansion positions is restricted by templates in \S \ref{sec:entity_elaboration}, we do not apply Filter 7 to the candidates derived by this approach.

\begin{table}[!t]
    \centering
    \small
    \begin{tabular}{p{7.0cm}}
    \toprule
    1. Is $X$ or $Y$ an obviously incoherent text whose PPL (\S \ref{paragraph:language_fluency})
    exceeds a threshold (e.g., 200)?
    \\\\
    2. Does any expanded modifier of $Y$ not contain any meaningful non-stopwords? 
    \\\\
    3. Is $X$ too short (e.g, less than 3 tokens)?
    \\\\
    4. Is any expanded modifier of $Y$ too long (e.g, more than 20 tokens)?
    \\\\
    5. Does any expanded modifier of $Y$ contain more than {3} consecutive punctuations or any special symbols such as ``http'', ``@'', etc.?
    \\\\
    6. Is the total length of expanded modifiers in $Y$ between 3 and 20 tokens?
    \\\\
    7. Does $Y$ only expand in one position?\\
    \\
    8. Does $Y$ express semantics inconsistent with $X$ where the Nli-E score (\S \ref{paragraph: semantic_fidelity}) is lower than a threshold (e.g., 0.5)?\\
    \bottomrule      
    \end{tabular}
    \caption{Filters applied to candidate pair $\{X, Y\}$.}
    \label{tab:filter_candidate}
\end{table}

\section{Data Statistics of Different Methods}\label{sec: detailed_data_statistics}
\begin{table}[!htbp]
    \centering
    \small
    \begin{tabular}{c c c c c c}
    \toprule
        & {NSC} & {CTP} & {MMP} & {IAR} & {Total} \\
        \midrule
        English & 5.0M & 5.0M & 1.5M & 0.5M & 12M\\
        Chinese & - & 5.5M & 5.5M & 1M & 12M\\
    \bottomrule
    \end{tabular}
    \caption{Statistics of the automatically constructed corpora. NSC, CTP, IAR and MMP are short for the approaches introduced in \S \ref{sec:sentence_compression}$\sim$\S \ref{sec:mask_prediction} respectively.}
    \label{tab:training_data_statistics}
\end{table}

\section{Details for Acquiring the Reference}\label{sec: refinement_guide}
We paid 2000 dollars for 5 experienced native-speaker annotators from the crowdsourcing platform for annotating the English and Chinese references, respectively. 
For each candidate pair $\{X, Y\}$ sampled from the automatically constructed corpus, in order to obtain a high-quality reference, we consider several quality dimensions while refining the expansion $Y$ according to the criteria in Table~\ref{tab:annotation_guidelines}. 
Since Fid. and Fert. are objective criteria, they are easy to be satisfied through simple revision. 
For the rest criteria, since they are more subjective for judgment, we provide 
some examples in Table~\ref{tab: evaluation_guide}
to show different grades of them.
After revision, we require the resulting reference expansion to be exactly fluent and coherent, and at least moderately informative and elegant. 
That is, the Flu. and Coh. scores must reach the full scale at 5, and the Info. and Eleg. scores should be at least 3.



\begin{table}[!t]
    \centering
    \small
    \begin{tabular}{p{7.0cm}}
    \toprule
    \textbf{Fidelity (Fid.)} : all the tokens from $X$ should appear in $Y$ and their relative order should remain the same. 
    \\\\
    \textbf{Fertility (Fert.):} $Y$ should expand at as many positions of $X$ as possible (at least 2 positions) and the overall expanded length should not be too short (at least 5 tokens).
    \\\\
    \textbf{Fluency (Flu.):} $Y$ should be a fluent text.
    \\\\
    \textbf{Coherence (Coh.):} each expanded modifier of $Y$ should be semantically coherent with its context and adherent to the central idea of $X$. 
    \\\\
    \textbf{Informativeness (Info.):}  the expanded modifiers of $Y$ should be rich in context rather than universal or dull.
    \\\\
    \textbf{Elegance (Eleg.): } It's better for the expanded modifiers to be elegantly worded. 
    \\
    \bottomrule
    \end{tabular}
    \caption{Criteria on various dimensions for the high-quality reference expansion $Y$, given the source text $X$.}
    \label{tab:annotation_guidelines}
\end{table}

\section{Examples with Low Nli-E Scores}\label{sec: incoherent_expansions}
We provide some semantically incoherent expansions in Table~\ref{tab:nli_example}, which are correctly assigned with low Nli-E scores. 
\begin{table}[!htbp]
    \centering
    \small
    \begin{tabular}{p{5.5cm}l c}
    \toprule
    \textbf{Expansions} & \textbf{Nli-E}\\
    \midrule
     if you are \textcolor[HTML]{f72585}{[\textit{no longer}]} sure of the exact date of your departure , [\textit{please}] do let me know [\textit{through the contact us page}] . & 0.0068\\
     \\
     \textcolor[HTML]{f72585}{[\textit{if}]} i had a glass of dom perignon at my wedding in august [\textit{, i would have loved it}] . & 0.0093 \\
     \\
     \textcolor[HTML]{f72585}{[\textit{double-check to make sure}]} your message has been sent successfully. & 0.2886 \\
     \bottomrule
    \end{tabular}
    \caption{Example incoherent expansions with low Nli-E scores. The [\textit{bracketed italic parts}] are expanded modifiers while the rest are identical to the source text. The \textcolor[HTML]{f72585}{red spans} are the triggers for semantic incoherence.}
    \label{tab:nli_example}
\end{table}

\section{Implementation Details for Automatic Metrics}\label{sec: implementation_details_metrics}
We utilize the outcomes of neural models to compute part of the automatic evaluation metrics. 
The details for these models are listed as follows. 
\begin{itemize}[wide=0\parindent,noitemsep, topsep=0.5pt]
\item \textbf{Fluency (PPL)} We use GPT2-base \footnote{English GPT2 model: \url{https://huggingface.co/gpt2}, Chinese GPT2 model: \url{https://huggingface.co/uer/gpt2-chinese-cluecorpussmall}.} to compute expansion fluency for both English and Chinese. 
\item \textbf{Coherence (Nli-E)} We use open-source NLI models \footnote{
English NLI model: \url{https://huggingface.co/ynie/roberta-large-snli\_mnli\_fever\_anli\_R1\_R2\_R3-nli}, Chiese NLI model: \url{https://huggingface.co/Jaren/chinese-roberta-wwm-ext-large-NLI}.} to estimate semantic coherence between the expansion and the source text. 
\item \textbf{Informativeness (Info-Gain)} The text-infilling model for computing Inverse-PPL and Inherent-PPL is initiated from a pre-trained T5 model \footnote{English T5 model: \url{https://huggingface.co/t5-base}, Chinese T5 model: \url{https://huggingface.co/uer/t5-base-chinese-cluecorpussmall}.} and further trained on our constructed TE corpus. 
During training, the input is the expansion where the spans from the source text are masked, and the output is the counterpart where the expanded modifiers are masked instead. 
\end{itemize}

\section{Implementation Details for Baseline Models}\label{sec: implementation_details_baselines}
We implement all the models using the base version of pre-trained model backbones from Hugging Face\footnote{\url{https://huggingface.co/models}}. 
We adopt the AdamW optimizer with a learning rate of 5e-5 and a weight decay of 0.01. 
All the models are trained on 8 V100 GPUs with a batch size of 512 for maximum of 20,0000 steps.

\section{Details for Human Evaluation}\label{sec: evaluation_guide}
We judge the Fluency (Flu.), Coherence (Coh.), Informativeness (Info.), and Elegance (Eleg.) of an expansion based on a 1-5 Likert scale, with 1 indicating unacceptable, 3 indicating moderate, 5 indicating excellent performance, 2 and 4 for handling unsure cases. 
Examples to show different grades on each dimension are provided in Table~\ref{tab: evaluation_guide}. 
The scores are further normalized into $[0, 100]$.
We randomly choose 100 samples from the Chinese TE test set. 
Three experienced Chinese annotators are recruited from the crowdsourcing platform and the average scores among them are reported.

\begin{table*}[!t]
    \centering
    \small
    \begin{tabular}{p{16cm}}
    \toprule
    \textbf{Fluency (Flu.)}
    \begin{itemize}
        \item[1 =] The expansion is totally unreadable, which does not make sense, and is obviously not a human-written text.
        \item[3 =] The expansion is readable but contains a few grammatical mistakes. 
        \item[5 =] The expansion is grammatically correct and the meaning is clearly understandable. 
    \end{itemize}
    \textbf{Coherence (Coh.)} 
    \begin{itemize}
        \item[1 =] The expansion conveys an idea that contradicts the central idea of the source text. 
        \begin{itemize}
            \item my \textcolor[HTML]{f72585}{[\textit{least}]} favorite sport is basketball
            \item my favorite sport is \textcolor[HTML]{f72585}{[\textit{football rather than}]} basketball
        \end{itemize}

        \item[3 =] The expansion subtly changes the degree of certainty or the scope of the fact described in the source text. 
        \begin{itemize}
            \item my favorite sport is \textcolor[HTML]{f72585}{[\textit{perhaps}]} basketball
            \item \textcolor[HTML]{f72585}{[\textit{in summer ,}]} my favorite sport is basketball \textcolor[HTML]{f72585}{[\textit{, but in winter , my favorite is skiing}]}
        \end{itemize}

        \item[5 =] The expansion accurately adds modifiers that are all coherent with the core semantics of the source text.
        \begin{itemize}
            \item my \textcolor[HTML]{3a86ff}{[\textit{personal}]} favorite sport \textcolor[HTML]{3a86ff}{[\textit{to play}]} is basketball \textcolor[HTML]{3a86ff}{[\textit{, since i was a little child}]}

            \item \textcolor[HTML]{3a86ff}{[\textit{there is no doubt that}]} my \textcolor[HTML]{3a86ff}{[\textit{all-time}]} favorite sport is basketball \textcolor[HTML]{3a86ff}{[\textit{, and i want to watch nba games live one day}]}
        \end{itemize}

    \end{itemize}
    \textbf{Informativeness (Info.)} \\
    \begin{itemize}
        \item[1 =] All the inserted modifiers in the expansion are universal and dull, or just repeat the content in the source text. 
        \begin{itemize}
            \item  \textcolor[HTML]{f72585}{[\textit{by the way}]} my favorite sport \textcolor[HTML]{f72585}{[\textit{, as you know ,}]} is basketball

            \item \textcolor[HTML]{f72585}{[\textit{she said that , ``}]} my favorite sport is \textcolor[HTML]{f72585}{[\textit{also}]} basketball \textcolor[HTML]{f72585}{[\textit{''}]}

            \item my favorite sport is basketball \textcolor[HTML]{f72585}{[\textit{, which is my favorite sport}]}
        \end{itemize}
        
        \item[3 =] 
        Only one inserted modifier is relevant to the context while the rest modifiers are still trivial.
        \begin{itemize}
            \item my \textcolor[HTML]{f72585}{[\textit{very}]} favorite sport \textcolor[HTML]{3a86ff}{[\textit{to watch}]} is basketball \textcolor[HTML]{f72585}{[\textit{, especially basketball}]}

           \item  \textcolor[HTML]{f72585}{[\textit{i think this is a good point ,}]} my favorite sport  \textcolor[HTML]{3a86ff}{[\textit{since childhood}]} is basketball  \textcolor[HTML]{f72585}{[\textit{( and i love it )}]}.
            
        \end{itemize}

        \item[5 =] At least two expanded modifiers are specific to the context of the source text and provide some in-depth information. 
        
        \begin{itemize}
            \item  \textcolor[HTML]{3a86ff}{[\textit{when it comes to sports ,}]} my \textcolor[HTML]{3a86ff}{[\textit{absolute}]} favorite sport \textcolor[HTML]{3a86ff}{[\textit{of all time}]} is basketball \textcolor[HTML]{3a86ff}{[\textit{, and i ’m a huge fan of james}]}

            \item \textcolor[HTML]{3a86ff}{[\textit{besides tennis ,}]} my \textcolor[HTML]{3a86ff}{[\textit{personal}]} favorite sport is basketball \textcolor[HTML]{3a86ff}{[\textit{, especially when i was in college}]}
        \end{itemize}

    \end{itemize}
    \textbf{Elegance (Eleg.)} \\
    \begin{itemize}
        \item[1 =] The expanded parts only contain meaningless gibberish, punctuation marks, or high-frequency stopwords.
        \begin{itemize}
            \item my \textcolor[HTML]{f72585}{[\textit{very}]} favorite sport is basketball \textcolor[HTML]{f72585}{[\textit{, too}]}

            \item \textcolor[HTML]{f72585}{[\textit{anyway ,}]} my favorite sport \textcolor[HTML]{f72585}{[\textit{now}]} is \textcolor[HTML]{f72585}{[\textit{actually}]} basketball    
        \end{itemize}

        \item[3 =] Only one expanded modifier is elegantly worded, which contains some graceful adjectives/adverbials, or high-level idioms, or vivid depictions, etc. 
        \begin{itemize}
            \item \textcolor[HTML]{f72585}{[\textit{so far}]} my favorite sport is basketball \textcolor[HTML]{3a86ff}{[\textit{, which i play for fun at all times .}]}

            \item \textcolor[HTML]{f72585}{[\textit{and of course ,}]} my \textcolor[HTML]{3a86ff}{[\textit{all-time}]} favorite sport \textcolor[HTML]{f72585}{[\textit{( except racquetball )}]} is basketball
        \end{itemize}

        \item[5 =] At least two expanded modifiers are considered elegantly worded.
        \begin{itemize}
            \item \textcolor[HTML]{3a86ff}{[\textit{as a sportsperson ,}]} my \textcolor[HTML]{3a86ff}{[\textit{personal}]} favorite sport is basketball \textcolor[HTML]{3a86ff}{[\textit{, and this has always been my passion .}]}

            \item my favorite sport \textcolor[HTML]{3a86ff}{[\textit{in the world}]} is basketball \textcolor[HTML]{3a86ff}{[\textit{, which i play for fun at all times .}]}
        \end{itemize}











        
    \end{itemize}\\
    \bottomrule
    \end{tabular}
    \caption{Evaluation guide on different quality dimensions based on a 1-5 Likert scale (1 for unacceptable, 3 for moderate, 5 for excellent 2 and 4 for unsure). Modifiers that are eligible under specific criteria are marked in \textcolor[HTML]{3a86ff}{bleu}, while the ineligible modifiers are marked in \textcolor[HTML]{f72585}{red}.}
    \label{tab: evaluation_guide}
\end{table*}

\end{CJK*}
\end{document}